\pdfoutput=1

\documentclass[11pt]{article}

\usepackage[]{EMNLP2023}

\usepackage{times}
\usepackage{latexsym}

\usepackage[T1]{fontenc}

\usepackage[utf8]{inputenc}

\usepackage{microtype}

\usepackage{inconsolata}

\usepackage{refstyle}

\usepackage{booktabs}
\usepackage{multirow} 
\usepackage{soul}
\usepackage{tabularx}
\usepackage{xspace}
\usepackage{color, colortbl}
\usepackage{colortbl}
\usepackage{makecell}

\usepackage{amsmath}
\DeclareMathOperator*{\argmax}{argmax} 
\usepackage{pifont}
\usepackage{amsfonts}
\usepackage{bm}

\usepackage{graphicx}
\usepackage[nodisplayskipstretch]{setspace}

\usepackage{cleveref}
\crefformat{section}{\S#2#1#3}
\crefformat{subsection}{\S#2#1#3}
\crefformat{subsubsection}{\S#2#1#3}
\crefrangeformat{section}{\S#3#1#4 to~\S#5#2#6}
\crefmultiformat{section}{\S#2#1#3}{ and~\S#2#1#3}{, #2#1#3}{ and~#2#1#3}
\Crefformat{figure}{#2Fig.~#1#3}
\Crefmultiformat{figure}{Figs.~#2#1#3}{ and~#2#1#3}{, #2#1#3}{ and~#2#1#3}
\Crefformat{table}{#2Tab.~#1#3}
\Crefmultiformat{table}{Tabs.~#2#1#3}{ and~#2#1#3}{, #2#1#3}{ and~#2#1#3}
\Crefformat{appendix}{#2Appx.~\S#1#3}
\crefformat{algorithm}{Alg.~#2#1#3}
\Crefformat{equation}{#2Eq.~#1#3}

\newcommand{\stitle}[1]{\vspace{1ex} \noindent{\bf #1.}}
\newcommand{\muhao}[1]{{\color{blue}[{MC:} #1]}}

\newcommand{\modelnamens}{\textsc{IoI}}
\newcommand{\modelname}{\modelnamens\xspace}

\title{Privacy-Preserving Language Model Inference with Instance Obfuscation}

\author{
Yixiang Yao\textsuperscript{1}, Fei Wang\textsuperscript{1}, Srivatsan Ravi\textsuperscript{1}, Muhao Chen\textsuperscript{2} \\
\textsuperscript{1}University of Southern California \\
\textsuperscript{2}University of California, Davis \\
\textsuperscript{1}\texttt{\{yixiangy,fwang598,srivatsr\}@usc.edu} \\
\textsuperscript{2}\texttt{muhchen@ucdavis.edu}
}

\begin{document}
\maketitle

\begin{abstract}

Language Models as a Service (LMaaS) offers convenient access for developers and researchers to perform inference using pre-trained language models. Nonetheless, the input data and the inference results containing private information are exposed as plaintext during the service call, leading to privacy issues. 
Recent studies have started tackling the privacy issue by transforming input data into privacy-preserving representation from the user-end with the techniques such as noise addition and content perturbation, while the exploration of inference result protection, namely \emph{decision privacy}, is still a blank page.
In order to maintain the black-box manner of LMaaS, conducting data privacy protection, especially for the decision, is a challenging task because the process has to be seamless to the models and accompanied by limited communication and computation overhead.
We thus propose Instance-Obfuscated Inference (\modelname) method, which focuses on addressing the decision privacy issue of natural language understanding tasks in their complete life-cycle. 
Besides, we conduct comprehensive experiments to evaluate the performance as well as the privacy-protection strength of the proposed method on various benchmarking tasks.

\end{abstract}
\section{Introduction}

Language Models as a Service (LMaaS; \citealt{sun2022black,brown2020language}) empowers researchers and developers to access pre-trained language models (PLMs) through cloud services without worrying about the complexities of model training, deployment, and infrastructure management. To interact with LMaaS, users usually send 
API
requests to the designated endpoints designed by the service providers and receive responses generated by the remote language models. Such a setup benefits both parties: on the one hand, users can jump-start on integrating the powerful PLMs into their data processing tasks; on the other hand, the underlying models and the processing pipelines, as the intellectual properties, are hidden from end users so that the service providers can protect them from leakage. 
However, given the lack of user control over the blackbox cloud service, the data in the requests can be illegally used by the service providers or potential attackers, thus causing privacy issues, including data leakage, unauthorized data access, profiling, and tracking~\cite{sen2015security,tang2016ensuring}.

\begin{figure}[!t]
    \centering
    \includegraphics[width=1\linewidth]{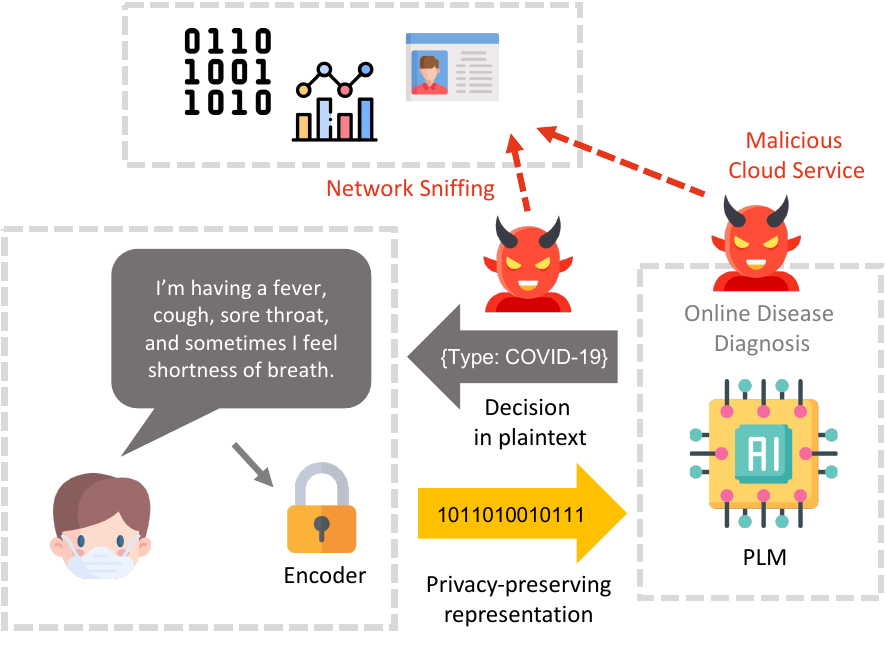}
    \vspace{-.8em}
    \caption{A privacy adversary example with state-of-the-art privacy protection in LMaaS. Despite encoding the end user's input into privacy-preserving representations, the raw output representations or decisions are still in plaintext, making them vulnerable to attacks from both network channels and servers.}
    \label{fig:example}
    \vspace{-1em}
\end{figure}

Recent literature has started to address the privacy issues of user inputs in LMaaS, for which solutions are typically based on techniques privatizing the input representation into intermediate ones. 
Methods of such kind include noise injection~\cite{plant-etal-2021-cape}, differential privacy (DP)~\cite{hoory-etal-2021-learning-evaluating,yue-etal-2021-differential,xu-etal-2020-differentially}, and adversarial training~\cite{li-etal-2018-towards,coavoux-etal-2018-privacy}.
Moreover, the intermediate representations are further fused or manipulated to prevent reverse engineering, while still remaining sufficient information for effective model inference~\cite{zhou-etal-2022-textfusion}. 
Unfortunately, to the best of our knowledge, none of the existing methods takes decision privacy into consideration, that is, the inferencing results are not protected which could implicitly or explicitly reveal users' sensitive information based on the specific tasks applied \cite{shejwalkar2021membership,kahla2022label}. For example, as shown in \Cref{fig:example}, a PLM employed by the online disease diagnosis service can analyze and determine the type of diseases based on the symptom descriptions from the patients. Even though privacy-preserving representations as the input can somehow protect the patients' submitted content, sensitive information such as the distribution of diseases \cite{mao2011loose} from the output aggregation still discloses to the malicious cloud service providers or the hackers via network sniffing.\footnote{The security of the network channel is not the scope of this paper. You can assume it is already end-to-end encrypted.}

Considering the significance and necessity of decision privacy protection, we propose to investigate a method that ensures the protection of both raw input content and raw output representation. However, protection in the decision phase could be more challenging than its counterpart in the input phase due to several reasons. First, unlike user inputs, since the final decision is made by the PLM on the cloud, users will have no direct means to intervene in it. 
Second, due to the required anonymity, incurred communication costs inevitably increase. Third, from the perspective of intellectual property protection, it is not practical for LMaaS providers to 
disclose parameters and architectures of the models, including the last few layers that are close to the decisions, to users.
These challenges call upon a solution that effectively protects the models' decisions before they come out, while does not violate the black-box nature of the LMaaS.

In this paper, we propose \modelname (\textsc{\underline{I}}nstance-\textsc{\underline{o}}bfuscated \textsc{\underline{I}}nference), which aims to protect the privacy of PLM decisions without losing the compatibility of utilizing the state-of-the-art input privacy protection approaches at inference phase. During inference, \modelname intentionally obfuscates the instance, hiding the raw decision distribution from revealing any sensitive information. However, the user who applies the obfuscation retains the ability to recover the true decision distribution. Note as a pilot study, \modelname focuses on text classification tasks.

\begin{figure}[!t]
    \centering
    \includegraphics[width=1\linewidth]{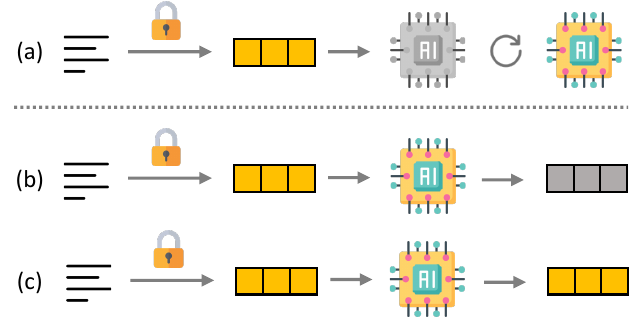}
    \vspace{-.8em}
    \caption{Privacy-preserving scenario comparison. \textbf{(a) Training Privacy} aims to protect the private training data. A typical privacy tool for this scenario is differential privacy. \textbf{Inference Privacy} includes \textbf{(b) Input Privacy} that prevents the raw input data from being revealed; and \textbf{(c) Decision Privacy} that protects the inference results. The vectors in \textcolor{orange}{orange} are privacy-preserving, while the ones in \textcolor{gray}{gray} are not.}
    \label{fig:method-comparison}
    \vspace{-1em}
\end{figure}

Despite distinctiveness, to avoid the ambiguity of understanding different privacy techniques, in \Cref{fig:method-comparison}, we summarize them according to the application scenarios. Specifically, SOTA methods utilize DP for training time data privacy. Other aforementioned noise addition or perturbation methods safeguard the raw input from being reverse-engineered. On the contrary, \modelname ensures the confidentiality of decisions from the model inference. 

The contributions of this work are three-folds. First, to the best of our knowledge, this is the first approach to explore the feasibility of protecting PLM \emph{decisions} in a black-box manner. Second, we define decision privacy, and comprehensively study the instance obfuscation strategies and privacy-preserving decision resolution in the context of it. Third, we define evaluation metrics for decision privacy, and empirically verify the performance and privacy strength of the proposed method.


\section{Privacy-Preserving Inference}
\label{sec:preliminary}

For a text classification task $M: \mathcal{X} \rightarrow \mathcal{Y}$, where $\mathcal{X}$ is the input text and $\mathcal{Y}$ is the label set. The \textit{privacy-preserving inference} takes a step further 
avoiding the exposure of any private information about the inputs and model decisions to the service provider.
While the encoding methods\footnote{The encoding mentioned in this paper is not by the PLM's encoder but as ``encryption''.} for protecting the privacy of $\mathcal{X}$ start emerging, the counterpart for $\mathcal{Y}$, which we call decision privacy, remains uncharted.

The intuition for achieving decision privacy is to make the model's raw decision as random as possible to all parties except the input instance owner, and the raw decision can only be recovered via a certain resolution method by the input instance owner. In the rest of this section, we formally define decision privacy in the context of text classification, as well as privacy-preserving inference.

\subsection{Decision Privacy}

For text classification, suppose $(\bm{x},y)$ is an instance of $(\mathcal{X},\mathcal{Y})$, and a finite label set $C = \{ c_i | 1 \leq i \leq n, n \geq 2 \}$ is the range of $\mathcal{Y}$. We say $M$'s output has \textit{perfect privacy} if

\begin{equation}
\label{equ:perfect-privacy}
    Pr[M(\bm{x}) = c_i] \approx \frac{1}{n},
\end{equation}

\noindent
that is, the probability of an adversary acquiring the predicted label $c_i$ from $M$ for the given input $\bm{x}$ is almost no better than a random guess.

However, directly adhering to \Cref{equ:perfect-privacy} leads to compromised functionality of $M$, since $M$ is essentially a random choice function and useless in practice. Instead, a certain encoding function $E(\cdot)$ can be performed on the input $\bm{x}$, so that the decision privacy of an arbitrary model $M$ is ensured by $E(\cdot)$: 

\begin{equation}
\label{equ:practical-privacy}
    | Pr[M(E(\bm{x})) = c_i] - \frac{1}{n} | \leq \epsilon,
\end{equation}

\noindent
where $\epsilon \in [0, 1)$ is seen as a privacy budget. Adjusting $\epsilon$ balances the utility and privacy: the smaller the $\epsilon$ is, the better the decision privacy. 


\subsection{Problem Definition} 

The privacy-preserving inference is defined as:

\begin{equation}
\label{equ:problem-definition}
    M(E(\bm{x})) \rightarrow y',
\end{equation}

\noindent
where the encoding function $E(\cdot)$ has \textbf{two} functionalities: (1) It encodes the raw $\bm{x}$ into some privacy-preserving representation remaining interpretable by $M$, which is already studied by previous work \cite{qu2021natural,yue-etal-2021-differential,zhou-etal-2022-textfusion} and is \textbf{not} the focus of this paper.
(2) The inference result transitions from the actual prediction $y$ to the privacy-preserving $y'$, whose distribution satisfies decision privacy defined by \Cref{equ:practical-privacy}:

\begin{equation}
\label{equ:decision-privacy-distribution}
    | Pr[y'=c_i] - \frac{1}{n} | \leq \epsilon.
\end{equation}

The privacy property is ensured by $E(\cdot)$. It is mathematically hard or impossible to find its inverse function $E^{-1}(\cdot)$, so that the adversary can not either recover the raw input $\bm{x}$ from the privacy-preserving representation $E(\bm{x})$, or the actual prediction $y$ from the privacy-preserving prediction $y'$. 
A decoding\footnote{This is different from the traditional ``decryption'' in information security.} function $D(\cdot)$ is available to decode true $y$ from $y'$ with the knowledge of the raw input and encoding settings, that is,

\begin{equation}
\label{equ:decoding}
    y \leftarrow D(y', E, \bm{x}).
\end{equation}

\Cref{equ:decoding} can be extended to the case that resolving a $y$ depends on mulitple $y'$s:

\begin{equation}
\label{equ:decoding-multi-y}
    y \leftarrow D(y_0', \cdots, y_g', E, \bm{x}),
\end{equation}

\noindent
where $y_0' \cdots y_g'$ are all necessary $y'$s to decode $y$. Note $E$ and $\bm{x}$ are for identifying these $y'$s, and could be omitted if the user maintains the reference between $y'$s and $y$.

Therefore, privacy-preserving inference allows the user to query LMaaS without exposing sensitive information to the service provider or the adversary, by sending $E(\bm{x})$ to the server and decoding $y$ from $y'$ with $D(\cdot)$ locally. Throughout the process, the adversary learns nothing from the encoded input $E(\bm{x})$ or encoded decision $y'$.

\stitle{Distinctions to DP or input privacy} In general, DP adds proper noise to the given input instances so that the individual information of input instances will not be leaked, but the overall statistical features of them remain. Similarly, most input privacy methods perturb the raw input to prevent input reverse-engineering while keeping the necessary information for inference. Hence, an ideal DP or input privacy method should satisfy $M(E(\bm{x}))\approx M(\bm{x})$, where $E$ is the corresponding DP or input privacy method, while protecting the privacy of $\bm{x}$. On the contrary, decision privacy tends to make $M(E(\bm{x}))$ as random as possible whereas $D(M(E(\bm{x}))) \approx M(\bm{x})$.

\section{Method}
\label{sec:method}
\vspace{-0.5em}

This section begins with an overview of our privacy-preserving inference framework for text classification. It follows by detailing the core component $E(\cdot)$ for encoding in \Cref{sec:instance-obfuscation}, and $D(\cdot)$ for decoding in \Cref{sec:decision-resolution}. Finally, analyses and justifications regarding privacy properties are provided in \Cref{sec:privacy-analysis}.

\begin{figure*}[t!]
    \centering
    \includegraphics[width=1\linewidth]{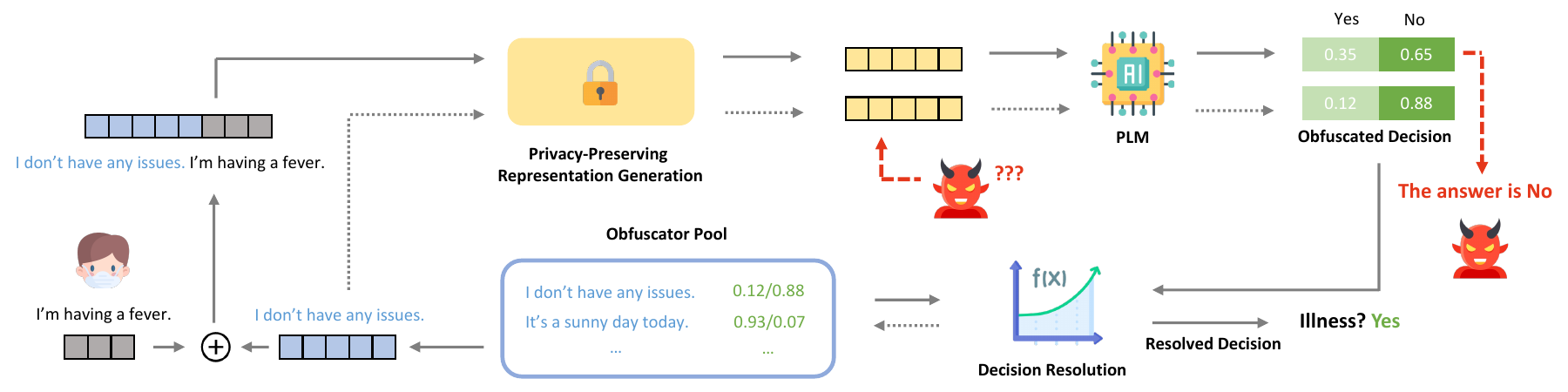}
    \vspace{-1.5em}
    \caption{The demonstration of \modelname workflow for decision privacy protection. If a user (bottom left) makes illness inquiries via a PLM-driven online diagnosis system, normally, the inference result will be returned in plain text.
    As a most basic example, in \modelname, the raw text is concatenated with an obfuscator, which is also a text. Subsequently, the concatenated text and the obfuscator are encoded respectively by the privacy-preserving representation generation module, which ensures the produced embedding representation is privacy-preserving (irreversible and unique). Consequently, instead of receiving one ``plaintext'', the PLM receives two independent ``ciphertext'' and makes inferences on them without knowing their correlation, raw text, and true decision.
    However, only the user is able to recover the true decision by leveraging the distribution of these two inferences. 
    In practice, each input text is obfuscated by a group of obfuscators, and the requests from multiple inputs are sent to the PLM in arbitrary order.
    }
    \label{fig:architure}
    \vspace{-1em}
\end{figure*}

The intuition behind \modelname is to obfuscate the raw instance with obfuscators so that the PLM's inference distribution is intentionally steered. Thus, the adversaries cannot deduce the true decision unless they possess knowledge of the corresponding resolution method and parameters.
The general workflow of \modelname is shown in \Cref{fig:architure}, consisting of instance obfuscation as $E(\cdot)$ and decision resolution as $D(\cdot)$. Instead of sending the raw instance $x$ to the PLM and acquiring the decision, \modelname conceals $x$ by concatenating it with an obfuscator $b$, which is also a text sequence (\Cref{sec:instance-obfuscation}). The concatenated text $[b;x]$ along with the obfuscator $b$ are sent to the privacy-preserving representation generation (PPRG) module, respectively, where the input is encoded by a compatible SOTA input privacy method. PPRG produces privacy-preserving representations, which are \underline{irreversible} and remain \underline{distinct} even for identical inputs, and are treated as inputs for the PLM.
After PLM's inference on PPRG-encode $b$ and $[b;x]$, the raw decision distribution of $[b;x]$ does not reflect the inference of $x$ since it is steered by the elaborated obfuscator $b$. But as the data owner, the actual decision $y$ can be resolved via decision resolution (\Cref{sec:decision-resolution}) by utilizing the decision distributions of $[b;x]$ and corresponding $b$.
We further show that the true $y$ is hard to be recovered from $y'$s in \Cref{sec:privacy-discussion}.

\subsection{Instance Obfuscation}
\label{sec:instance-obfuscation}

Sending the input instance $x$ in plaintext to the PLM reveals the input completely. Hence, some previous studies \cite{zhou-etal-2022-textfusion,plant-etal-2021-cape} employ a \emph{privacy-preserving text representation} that transforms the input into ``ciphertext'' form by perturbing representations. In this way, although the raw input content is not exposed, the output of the PLM still carries meaningful information and may be exploited by the adversary. To tackle the flaw of limited protection of the decision by previous methods, \modelname uses instance obfuscation, acting as $E(\cdot)$ in \Cref{sec:preliminary}. It not only protects the input privacy by reusing the existing SOTA privacy-preserving text representation methods as PPRG but also ``fools'' the adversary with baffled output for decision privacy.

The instance obfuscation is motivated by mixup~\cite{zhang2018mixup}, originally proposed for data augmentation. \citet{zhang2018mixup} theoretically shows that mixup can produce virtual feature-target pairs sampled from the same vicinal distribution as the original data. Specifically, it shows that, through a mapping (i.e., the LM in this context), the mixup of two raw inputs can be mapped to the mixup of their corresponding labels.
Based on that, if $E(\cdot)$ conceals the real instance $x$ by mixing it up with dummy instances, the PLM only makes an inference on the mixup instance without seeing $x$, and the proportion of dummy instances participated in the mixup steers the final decision. We call these dummy instances obfuscators; thus, $E(\cdot)$ obscures the true instance with selected obfuscators and let the PLM make decisions based on the elaborated input. The obfuscation process is the key for concealing information in a black-box LMaaS setting. This leads to the question of what is considered to be a high-quality obfuscator and how to obfuscate $x$ with $b$ properly, hence maximizing the performance as well as privacy protection.

\stitle{Obfuscator Selection} Obfuscators are simply normal (unlabeled) sentences that could be with or without any relation to the real instances to be protected. To be used as an obfuscator, an instance requires to have a corresponding predicted label from PLM. Note that the predicted label does not need to be correct so there is no need for a gold label. Thus, an obfuscator $b$ could be a sentence from any arbitrary corpus.

To steer the PLM's decision towards being affected by $b$ instead of $x$, we prioritize $b$ instances with higher confidence regarding the PLM decision. For example, in a binary classification task, if an instance $x_1$ scored 0.9 for label 1 and $x_2$ scored 0.7, then $x_1$ is picked over $x_2$ for $x_1$ is more deterministic to get label 1. Since the selected obfuscators can be paired with any real instances, an optimized way to re-use them is to have them pre-computed in an \textit{obfuscator pool}.


\stitle{Obfuscator Balancing} Based on the observation, a single $b$ instance for obfuscating $x$ results in the un-stableness in decision resolution (\Cref{sec:decision-resolution}) due to the uneven distribution of the PLM's decision. For example, in a 3-class classification, assume $b_1$ has label $c_1$, $b_2$ has label $c_2$, and $b_3$ has label $c_3$. After a single obfuscation with $b_1$, the label of $[b_1;x]$ predicted by $M$ could remain $c_1$, or change to $c_2$ or $c_3$. Thus, the steering of decision distribution using a single obfuscator is not steady. 
Balancing, as a solution, is employed to mitigate this issue. Specifically, each real instance $x$ is paired with \emph{at least} one \emph{unit group} of obfuscators. A unit group of obfuscators is defined as a set containing obfuscators with uniformly distributed labels from the label set $C$, that is,


\begin{equation}
\label{equ:obfuscator-unit-group}
    g=\{b_j | M(b_j) = c_i, b_j \in B, \forall c_i \in C\},
\end{equation}

\noindent
where $B$ is the obfuscator pool, and $|g|=|C|$. Moreover, a group can consist of more than one unit group. Formally, a group contains $n$ unit group is defined as

\begin{equation}
\label{equ:obfuscator-group}
    G_n = g_1 \cup g_2 \cup \cdots \cup g_n.
\end{equation}

Therefore, the obfuscated instances of $x$ are noted as $[b_i;x]$ where $b_i \in G_n$.
Using balancing in the previous example, $x$ concatenates with all three obfuscators and results in three obfuscated instances $[b_1;x]$, $[b_2;x]$ and $[b_3;x]$.

\stitle{Privacy-Preserving Representation Generation} Even though the raw instance $x$ is replaced with $[b;x]$ and $x$, the content remains in plaintext.
To protect their privacy, \modelname uses PPRG, being compatible with any compatible SOTA input privacy methods~\cite{zhou-etal-2022-textfusion,plant-etal-2021-cape}, for transforming $[b;x]$ and $b$ from text sequences to privacy-preserving representations. A qualified input privacy method has two requirements regarding privacy. First, the produced representation is not invertible so that the adversary can not reverse it back to plaintext. Second, the input privacy method is equipped with randomness so that the produced representation is distinct even for identical inputs.

After applying PPRG, the representations from multiple $x$s should be sent to the PLM in arbitrary order. This prevents the adversary from pairing up the encoded $[b;x]$ and $b$. A more detailed discussion regarding privacy is in \Cref{sec:privacy-discussion}.

\subsection{Privacy-Preserving Decision Resolution}
\label{sec:decision-resolution}

While the obfuscated instance ceases the raw instance $x$ from being accessible by the PLM, the true decision of $x$ is concealed in the concatenated result $y'$s as well. We outline a decision resolution method, as $D(\cdot)$ in \Cref{sec:preliminary} (\Cref{equ:decoding-multi-y}), to resolve true $y$ from multiple associated $y'$s.

As the balancing described in \Cref{sec:instance-obfuscation}, successfully executing $D(\cdot)$ to get the decision of $x$ requires all the associated $[b;x]$ and $b$ pairs. As to adversaries, correctly locating all associated instances from a tremendous amount of mixed instances, which are sufficiently obfuscated and randomized, is equivalent to finding a needle in a haystack. We have detailed analysis in \Cref{sec:privacy-discussion}.

Oppositely, as the data owner, running $D(\cdot)$ is as easy as pie. The strategy to separate $x$'s result $y$ from $y'$s is based on the divergence between the decision distribution of $[b;x]$ and $b$. Specifically, if $x$'s label is $c_k$, blending it with $b$ shifts the confidence of $[b;x]$'s decision distribution towards $c_k$ regardless of the $b$'s label. Taking our example in \Cref{fig:architure}, $[b;x]$ has 0.35 for ``yes'' and 0.65 for ``no,'' while $b$ has 0.12 for ``yes'' and 0.88 for ``no''. The confidence of ``yes'' for $[b;x]$ increases because of the involvement $x$, thus $x$ is highly like to be ``yes.''

Without loss of generality, we inductively evaluate such divergence over a unit group $g$, the decision label raises to be the label of $x$ if the confidence difference between the obfuscated instance and obfuscator that increases the most, that is,

\begin{equation}
\label{equ:decision-resolution-unit}
    \argmax_{1 \leq i \leq |C|} (c^{b_j;x}_i - c^{b_j}_i),
\end{equation}

\noindent
where $b_j \in g$, $c_i \in C$, $i$ denotes label id ($|C|$ is the number of labels) and $j$ denotes the obfuscator id. If the obfuscator group is extended to $G_n$, \Cref{equ:decision-resolution-unit} can be generalized as

\begin{equation}
\label{equ:decision-resolution}
    \argmax_{1 \leq i \leq |C|} \frac{1}{n}\sum_{j=1}^{|G_n|} (c^{b_j;x}_i - c^{b_j}_i),
\end{equation}

\noindent
where $b_j \in G_n$ and the confidence difference on the decision distribution is averaged.\footnote{Note that this succinct but effective strategy can seamlessly be applied to various privacy-preserving inference tasks, for it is not restricted by the number of labels.}

\section{Experiments}
\vspace{-0.5em}


We first introduce the datasets, baselines, and evaluation metrics in \Cref{sec:experimental-setup}. The main results regarding task performance and decision privacy are illustrated in \Cref{sec:main-results}. We further study the functionalities of technical components in \Cref{sec:analysis}.
\begin{table*}[t]
\centering
\setlength{\tabcolsep}{8pt}
\setlength\extrarowheight{0.5pt}
\small
\begin{tabular}{c|l|cc|ccc}
\hline\hline 
Dataset & Method & $T_r$ & $T_o$ & $\Phi_{r} \downarrow$ & $\Phi_{o} \downarrow$ & $\Phi \downarrow $ \\ 
\hline
\multirow{6}*{\shortstack[c]{SST-2\\(Acc.)}}
                     & \cellcolor{gray!20}Fine-tuned & \cellcolor{gray!20} .924 & \cellcolor{gray!20} .924 & \cellcolor{gray!20} $-$ & \cellcolor{gray!20} $-$ & \cellcolor{gray!20} $-$ \\
                     & \cellcolor{gray!20}Random Guess & \cellcolor{gray!20} .500 & \cellcolor{gray!20} .500 & \cellcolor{gray!20} $-$ & \cellcolor{gray!20} $-$ & \cellcolor{gray!20} $-$ \\ 
                     \cline{2-7}
                     & PP-BERT & .909 & .909 & .016 & .818 & .417 \\
                     & SanText+ & .830 & .830 & .102 & .660 & .381 \\
                     & TextFusion & .904 & .904 & .022 & .808 & .415  \\ 
                     & \modelname & .913 & .770 & \textbf{.012} & \textbf{.540} & \textbf{.276} \\
\hline
\multirow{6}*{\shortstack[c]{MRPC\\(Acc./F1)}}
                     & \cellcolor{gray!20}Fine-tuned & \cellcolor{gray!20} .860/.904 & \cellcolor{gray!20} .860/.904 & \cellcolor{gray!20} $-$ & \cellcolor{gray!20} $-$ & \cellcolor{gray!20} $-$ \\ 
                     & \cellcolor{gray!20}Random Guess & \cellcolor{gray!20} .500/.500 & \cellcolor{gray!20} .500/.500 & \cellcolor{gray!20} $-$ & \cellcolor{gray!20} $-$ & \cellcolor{gray!20} $-$ \\
                     \cline{2-7}
                     & PP-BERT & .434/.294 & .434/.294 & .489/.675 & \textbf{.132}/.412  & .310/.543 \\
                     & SanText+ & .711/.750 & .711/.750 & .164/.170 & .422/.500 & .293/.335 \\ 
                     & TextFusion & $-$/.882 & $-$/.882 & $-$/\textbf{.024} & $-$/.764 & $-$/.394 \\
                     & \modelname & .745/.794 & .570/.628 & \textbf{.124}/.122 & .166/\textbf{.256} & \textbf{.132}/\textbf{.189} \\
\hline
\multirow{5}*{\shortstack[c]{SST-5\\(Acc.)}}
                     & \cellcolor{gray!20}Fine-tuned & \cellcolor{gray!20} .500 & \cellcolor{gray!20} .500 & \cellcolor{gray!20} $-$ & \cellcolor{gray!20} $-$ & \cellcolor{gray!20} $-$ \\
                     & \cellcolor{gray!20}Random Guess & \cellcolor{gray!20} .200 & \cellcolor{gray!20} .200 & \cellcolor{gray!20} $-$ & \cellcolor{gray!20} $-$ & \cellcolor{gray!20} $-$ \\
                     \cline{2-7}
                     & PP-BERT & .490 & .490 & \textbf{.020} & .362 & .191  \\
                     & SanText+ & .426 & .426 & .148 & .282 & .215 \\
                     & \modelname & .467 & .339 & .066 & \textbf{.174} & \textbf{.120} \\
\hline
\multirow{5}*{\shortstack[c]{QNLI\\(Acc.)}}
                    & \cellcolor{gray!20}Fine-tuned & \cellcolor{gray!20} .915 & \cellcolor{gray!20} .915 & \cellcolor{gray!20} $-$ & \cellcolor{gray!20} $-$ & \cellcolor{gray!20} $-$ \\
                    & \cellcolor{gray!20}Random Guess & \cellcolor{gray!20} .500 & \cellcolor{gray!20} .500 & \cellcolor{gray!20} $-$ & \cellcolor{gray!20} $-$ & \cellcolor{gray!20} $-$ \\
                    \cline{2-7}
                    & PP-BERT & .658 & .658 & .281 & .316 & .298 \\
                    & SanText+ & .725 & .725 & .208 & .450 & .329 \\ 
                    & \modelname & .849 & .648 & \textbf{.072} & \textbf{.296} & \textbf{.184} \\
\hline\hline
\end{tabular}
\caption{Performance of resolved and obfuscated decisions by \modelname and baselines. $T_r$ indicates the task raw performance after decision resolution by the data owner, while $T_o$ indicates the performance that model owner or attacker retrieves. $\Phi_r$ and $\Phi_o$ measure how close $T_r$ and $T_o$ to the baseline and the random guess, respectively, and $\Phi$ is a balance between $\Phi_r$ and $\Phi_o$. The smaller the three $\Phi$s, the better the task performance and decision privacy protection.
The best result in each task is highlighted in \textbf{bold}. Only \modelname has effects on decision privacy, while the other baselines are either non-privacy-preserving or only protect input privacy.}
\label{tab:main-result}
\vspace{-1em}
\end{table*}

\subsection{Experimental Setup}
\label{sec:experimental-setup}

\stitle{Datasets} Our experiments are conducted on four benchmark datasets that span across various text classification tasks. \textbf{SST-2}~\cite{socher-etal-2013-recursive} requires to classify the sentiment of the given text into either positive or negative class.
\textbf{SST-5}, as an extension of the SST-2, granularizes the binary label into five categories: very negative, negative, neutral, positive, and very positive.
\textbf{MRPC}~\cite{dolan-brockett-2005-automatically} is a paraphrase identification task to determine whether two sentences 
are paraphrases.
\textbf{QNLI}~\cite{wang-etal-2018-glue}, derived from SQuAD~\cite{rajpurkar-etal-2016-squad}, 
is a natural language inference task seeking to identify if the context sentence contains the answer to the question.

\stitle{Baselines} Although no direct comparable methods regarding decision privacy are available, we select four reasonable and related baselines.
\textbf{Fine-tuned} is task fine-tuned model 
without privacy protection.
\textbf{Random Guess} 
denotes the random guess results.
\textbf{PP-BERT}~\cite{qu2021natural} is a privacy-preserving encoder that perturbs the token embeddings by adding random noise $N=rp$ where $r$ is the distance from the origin and $p$ is a unit hypersphere. $r$ is sampled from the Gamma distribution $\Gamma(n, \frac{1}{\eta})$.\footnote{We set $\eta=100$, a moderate value in the original paper, for balancing the noise strength and accuracy.} 
\textbf{SanText+}~\cite{yue-etal-2021-differential} replaces sensitive words with GloVe~\cite{pennington-etal-2014-glove} and utilizes differential privacy to ensure the privacy of the sanitized words.\footnote{Considering fairness, we set $\epsilon=3$ in SanText+ and use sanitized text directly in inference.} \textbf{TextFusion}~\cite{zhou-etal-2022-textfusion} alters the input text sequence or intermediate representations by eliminating redundant or sensitive information.
Note PP-BERT, SanText+, and TextFusion were all designed for input privacy.

\stitle{Metrics} Generally, the performance is evaluated by task-specific metrics (Accuracy/F1) denoted as $T$. To measure the raw performance, we use the obfuscated ($T_{o}$) for the obfuscated version of the decision, and the resolved ($T_{r}$) for the true (original) performance from the decision resolution.

Besides, to quantify the effectiveness of the decision privacy protection and decision resolution, we additionally define $\Phi_{r}=\frac{T_{\texttt{baseline}}-T_r}{T_{\texttt{baseline}}}$ and $\Phi_{o}=\frac{|T_o-T_{\texttt{random}}|}{1-T_{\texttt{random}}}$. They measure the relative performance difference from resolved $T_r$ to non-privacy-preserving baseline, and from obfuscated $T_o$ to random guess, respectively. Finally, a unified metric $\Phi = \frac{\Phi_{o}+\Phi_{r}}{2}$ measures the balance between the obfuscation strength and the task performance.\footnote{All three metrics are scaled to be in the range $[0,1]$, and the smaller value indicates better performance.}





\subsection{Main Results}
\label{sec:main-results}

\begin{table}[t]
\centering
\small
\setlength{\tabcolsep}{5pt}
\begin{tabular}{r|cccc}
\hline
& SST-2 & MRPC & SST-5 & QNLI \\
\hline
Max sequence len & 128 & 512 & 128 & 256 \\
Min confidence of $b$ & $>0.99$ & $>0.90$ & $>0.90$ & $>0.99$ \\
Group size $n$ & 1 & 1 & 1 & 1 \\
\hline
\end{tabular}
\caption{\modelname settings for main results}
\label{tab:ioi-settings}
\vspace{-.5em}
\end{table}

The parameter settings of \modelname are in \Cref{tab:ioi-settings}.\footnote{PPRG is not enabled in this experiment and the effect of it is studied in \Cref{sec:analysis}.}
The backbone PLM used in each task is fine-tuned and consistent with all baselines because PP-BERT, SanText+, TextFusion, and \modelname are applied in the inference phase.
As the main results presented in \Cref{tab:main-result}, \modelname performs almost the best among all tasks regarding the resolved ($\Phi_r$) and obfuscated ($\Phi_o$) results, and the balance between them ($\Phi$), except few are lower but close to the best baselines. Note that only \modelname can protect decision privacy, thus, its $\Phi_o$ is the best as compared to all others.

Specifically, on SST-2 and QNLI, the resolved results $T_r$ by \modelname have similar accuracy as the non-privacy fine-tuned baselines indicated by the smaller $\Phi_r$ while still deviating obfuscated result $T_o$ to be as close to random as possible showing as the smaller $\Phi_o$. 
For SST-5, as a harder version of SST-2, albeit it is not the best regarding task performance, \modelname balances the trade-off, indicated by $\Phi$, to still archive relatively better decision privacy.
For MRPC, the evaluation metrics capture the abnormal performance of PP-BERT, whose resolved prediction performance $T_r$ is worse than a random guess. In this case, while its $\Phi_o$ seems to be the best among all other methods, the high $\Phi_r$ and $\Phi$ precisely reflect its poor balance between task performance and decision privacy.

\subsection{Analyses}
\label{sec:analysis}

We further study the influence of the technical components described in \Cref{sec:method}.


\stitle{Obfuscator Selection} To verify the loose policy of obfuscator selection that any normal sentence can be a qualified obfuscator, we conduct the following contrastive experiment on SST-2 and SST-5. Specifically, we test the real instances with the obfuscator from the same and different datasets. As shown in \Cref{tab:ofuscator-selection}, using instances from a different dataset as obfuscators is indistinguishable from the ones from the same dataset.

\begin{table}[t]
\centering
\small
\setlength{\tabcolsep}{5pt}
\begin{tabular}{cccc}
\hline
Evaluation & Obfuscator & $T_r$ & $T_o$ \\
\hline
SST-2 & SST-2 & 0.907 & 0.770 \\
\rowcolor{gray!20}SST-2 & QNLI & 0.891 & 0.777 \\
SST-5 & SST-5 & 0.467 & 0.339 \\
\rowcolor{gray!20}SST-5 & QNLI & 0.461 & 0.331 \\
\hline
\end{tabular}
\caption{The performance impact with obfuscators from the same and different datasets}
\label{tab:ofuscator-selection}
\vspace{-.5em}
\end{table}

\begin{figure}[!t]
    \centering
    \includegraphics[width=.95\linewidth]{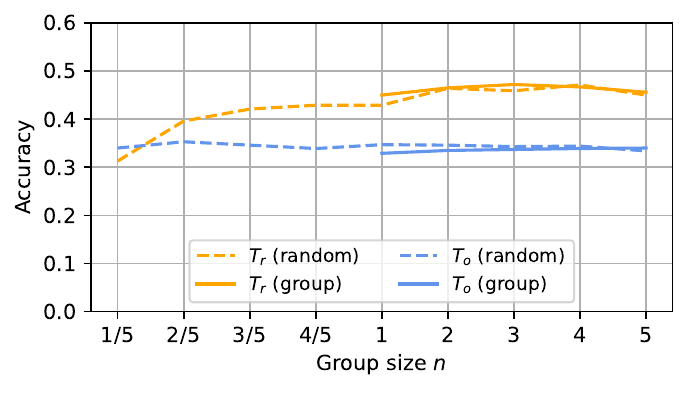}
     \vspace{-.5em}
    \caption{Balancing with different group size (SST-5)}
    \label{fig:balancing}
     \vspace{-.5em}
\end{figure}

\stitle{Balancing} This technique is intended for mitigating the issue of unbalanced obfuscator distributions, thus increasing the accuracy for decision resolution. 
Here, we study the necessity of balancing. Because SST-5 can be identified as a 5-class classification problem, a unit group $g$ (\Cref{equ:obfuscator-unit-group}), in which the obfuscator's labels are uniformly distributed, contains five obfuscators with different labels. According to \Cref{equ:obfuscator-group}, a group $G_n$ consists of $n$ $g$s. We set the $n$ to be from 1 to 5, that is, 5 to 25 obfuscators. Additionally, as the baseline, we test the obfuscation without balancing by randomly sampling 1 to 25 obfuscators from the obfuscator pool regardless of the classes they belong to. In \Cref{fig:balancing}, the performance of applying balancing is presented in a solid line, and the one for the randomly sampled obfuscators is in the dashed line. 
For the resolved version, without balancing, the accuracy $T_r$ (random) improves by more than 15\% from one randomly sampled obfuscator to five, and fluctuates relatively smooth after having more than a unit group of obfuscators (orange dashed line). With balancing $G_n$ ($T_r$, orange solid line), where $n$ ranges from 1 to 5, performs overall better and more steadily than the random samples. Unlike the resolved version, which receives the performance gains, for the obfuscated version, the performance remains stable with different $n$, and is outperformed by the resolved version for more than 10\% when the group size is at least a unit (blue lines). As a consequence, balancing archives the best and most robust performance for the resolved version, meanwhile maintaining the maximum gap to the obfuscated version for better decision privacy protection.


\stitle{Privacy-Preserving Representation Generation} PPRG utilizes the compatible input privacy method in a black-box fashion to transform obfuscated instances and obfuscators into representations that preserve privacy. Since the privacy strength and the ability to prevent attacks of the input privacy method are already comprehensively studied in its corresponding papers and follow-up works, we focus on the performance of plugging it with \modelname.

\Cref{tab:pprg} demonstrates the performance of \modelname when employing PP-BERT as PPRG. We test PP-BERT with various $\eta$ and compute its accuracy on SST-2. We then use it as PPRG and report $T_r$ and $T_o$. From the observation, the difference between the resolved result $T_r$ and PP-BERT is trivial (see $\%$), which indicates the strong compatibility and recovering ability of \modelname. Hence, we conclude that \modelname's task performance is dominated by the selected PPRG method as well as input privacy, meanwhile keeping the decision privacy.

\begin{table}[t]
\centering
\small
\setlength{\tabcolsep}{5pt}
\begin{tabular}{l|c|cc|c}
\hline
PP-BERT & Acc & $T_r$ & $T_o$ & \% \\
\hline
$\eta=200$ & .914 & .912 & .768 & .002 \\
\rowcolor{gray!20}$\eta=100$ & .925 & .908 & .759 & .018 \\
$\eta=50$ & .851 & .846 & .698 & .006 \\
\rowcolor{gray!20}$\eta=25$ & .536 & .528 & .516 & .015 \\
\hline
\end{tabular}
\caption{PPRG-enabled \modelname performance (SST-2)}
\label{tab:pprg}
\vspace{-1.8em}
\end{table}

\section{Related Work}
\vspace{-0.5em}

\stitle{Privacy Preservation in LMaaS} 
Recent studies are actively engaged in addressing the privacy concerns associated with LMaaS. Methods including noise injection~\cite{plant-etal-2021-cape}, differential privacy~\cite{hoory-etal-2021-learning-evaluating,yue-etal-2021-differential,xu-etal-2020-differentially}, and adversarial training~\cite{li-etal-2018-towards,coavoux-etal-2018-privacy}, and representation fusion \cite{zhou-etal-2022-textfusion} tend to perturb the input text sequence or intermediate representations by reducing unnecessary or sensitive information for PLM's inference. 
There also exist approaches~\cite{feng2020cryptogru,chen-etal-2022-x} that seek to protect data flow end-to-end, relying on homomorphic encryption,
albeit the execution of such models is very time-consuming and computationally expensive, and needs to modify the model from the server side.
On the other hand, to mitigate privacy issues in cloud PLM fine-tuning, offsite-tuning~\cite{xiao2023offsite} compresses the full PLM into a distilled version that allows users to tune plug-in adapters on their local, which protects the privacy of the user as well as the weights of PLM. \citet{du2023sanitizing} exploit local differential privacy to sanitize the embedding (and labels) for fine-tuning. 
However, none of the above work protects inference decision privacy of the LM under the black-box setting, which is exactly the focus of this work.


\stitle{Data Obfuscation} Although mixup~\cite{zhang2018mixup} is designed to alleviate the undesirable drawbacks of large deep neural networks, the concept of data combination and its effect on inference with minimal computation overhead is valuable and worth learning. \citet{guo2019augmenting} extend the mixup into the NLP world, 
and Co-mixup~\cite{kim2021co} discovers the possibility of applying mixup on multiple instances. 
Besides representation mixup, recent studies also obfuscate authorship of text by neutralizing the stylistic features of text with techniques, such as back-translation or representation disentanglement \cite{mahmood2022girl,altakrori-etal-2022-multifaceted,bevendorff-etal-2019-heuristic}.
Our instance obfuscated technique is inspired by representation mixup, while representing a pilot approach for LM decision protection.
\section{Conclusion}
\vspace{-0.5em}

In this work, we introduce decision privacy and propose \modelname that prohibits information leakage of PLMs under the settings of black-box LMaaS. In contrast to prior works, we are the first ones to consider the privacy protection of PLM's decision via instance obfuscation. Correspondingly, we define the evaluation metrics tailored for decision privacy and conduct comprehensive experiments regarding task performance and privacy protection. We anticipate our work conveys valuable insight and sheds some light on the trajectory of privacy in NLP.
\section*{Limitations}

We discuss two main limitations of this work. First, the extra instance inference cost. \modelname hides the target instance behind obfuscators so that the target instance never exposes directly to the PLM. To guarantee the strength of privacy protection and stability of the task performance, the strategies, including balancing and randomization, emit additional requests which result in multiple inferences for one instance. However, such incurred cost is not as severe as the previous works, for some of them need significant effort to fine-tune the remote PLMs, and some others require partial/entire model sharing hence compromising the privacy of the model.

Second, \modelname is deliberated for text classification tasks in a solely black-box fashion, thus it is not suitable for generative tasks. As for natural language generation, the adaption based on the current method requires addressing the problems such as mix-up tokens and variable lengths of generated text which are non-trivial. We leave this to be the future work.

\section*{Ethical Considerations}

Technology innovations generally offer potential benefits, but they also possess the risk of intentional exploitation for nefarious purposes, and LMaaS is not immune to this reality.
The presence of regulations and standards establishes a legal structure that ensures responsible utilization of data and grants individuals the right to request the deletion of their data. In the absence of such regulations, society depends on the ethical responsibility of technology practitioners to ensure the ethical usage of data. 

Decision privacy, defined in this paper, provides a fundamental direction for protecting the data, as well as LMaaS, from being abusively used. The proposed method technically guarantees the privacy of the input and output data to and from the LMaaS being fully obfuscated. Adopting this method ensures the operations and data are intended for legitimate purposes rather than malicious. Besides, the method itself can seamlessly be integrated into compatible underlying technologies or running systems without any extra modification, which reduces the barriers associated with implementation for increasing accessibility for individuals or organizations.

All experimental datasets used in this work are openly available benchmarks. No demographic or identity characteristics are used in this paper.

\bibliography{anthology,reference}

\begin{thebibliography}{31}
\expandafter\ifx\csname natexlab\endcsname\relax\def\natexlab#1{#1}\fi

\bibitem[{Altakrori et~al.(2022)Altakrori, Scialom, Fung, and
  Cheung}]{altakrori-etal-2022-multifaceted}
Malik Altakrori, Thomas Scialom, Benjamin C.~M. Fung, and Jackie Chi~Kit
  Cheung. 2022.
\newblock \href {https://aclanthology.org/2022.emnlp-main.153} {A multifaceted
  framework to evaluate evasion, content preservation, and misattribution in
  authorship obfuscation techniques}.
\newblock In \emph{Proceedings of the 2022 Conference on Empirical Methods in
  Natural Language Processing}, pages 2391--2406, Abu Dhabi, United Arab
  Emirates. Association for Computational Linguistics.

\bibitem[{Bevendorff et~al.(2019)Bevendorff, Potthast, Hagen, and
  Stein}]{bevendorff-etal-2019-heuristic}
Janek Bevendorff, Martin Potthast, Matthias Hagen, and Benno Stein. 2019.
\newblock \href {https://doi.org/10.18653/v1/P19-1104} {Heuristic authorship
  obfuscation}.
\newblock In \emph{Proceedings of the 57th Annual Meeting of the Association
  for Computational Linguistics}, pages 1098--1108, Florence, Italy.
  Association for Computational Linguistics.

\bibitem[{Brown et~al.(2020)Brown, Mann, Ryder, Subbiah, Kaplan, Dhariwal,
  Neelakantan, Shyam, Sastry, Askell et~al.}]{brown2020language}
Tom Brown, Benjamin Mann, Nick Ryder, Melanie Subbiah, Jared~D Kaplan, Prafulla
  Dhariwal, Arvind Neelakantan, Pranav Shyam, Girish Sastry, Amanda Askell,
  et~al. 2020.
\newblock Language models are few-shot learners.
\newblock \emph{Advances in neural information processing systems},
  33:1877--1901.

\bibitem[{Chen et~al.(2022)Chen, Bao, Huang, Dong, Jiao, Jiang, Zhou, Li, and
  Wei}]{chen-etal-2022-x}
Tianyu Chen, Hangbo Bao, Shaohan Huang, Li~Dong, Binxing Jiao, Daxin Jiang,
  Haoyi Zhou, Jianxin Li, and Furu Wei. 2022.
\newblock \href {https://doi.org/10.18653/v1/2022.findings-acl.277} {{THE}-{X}:
  Privacy-preserving transformer inference with homomorphic encryption}.
\newblock In \emph{Findings of the Association for Computational Linguistics:
  ACL 2022}, pages 3510--3520, Dublin, Ireland. Association for Computational
  Linguistics.

\bibitem[{Coavoux et~al.(2018)Coavoux, Narayan, and
  Cohen}]{coavoux-etal-2018-privacy}
Maximin Coavoux, Shashi Narayan, and Shay~B. Cohen. 2018.
\newblock \href {https://doi.org/10.18653/v1/D18-1001} {Privacy-preserving
  neural representations of text}.
\newblock In \emph{Proceedings of the 2018 Conference on Empirical Methods in
  Natural Language Processing}, pages 1--10, Brussels, Belgium. Association for
  Computational Linguistics.

\bibitem[{Dolan and Brockett(2005)}]{dolan-brockett-2005-automatically}
William~B. Dolan and Chris Brockett. 2005.
\newblock \href {https://aclanthology.org/I05-5002} {Automatically constructing
  a corpus of sentential paraphrases}.
\newblock In \emph{Proceedings of the Third International Workshop on
  Paraphrasing ({IWP}2005)}.

\bibitem[{Du et~al.(2023)Du, Yue, Chow, and Sun}]{du2023sanitizing}
Minxin Du, Xiang Yue, Sherman~SM Chow, and Huan Sun. 2023.
\newblock Sanitizing sentence embeddings (and labels) for local differential
  privacy.
\newblock In \emph{Proceedings of the ACM Web Conference 2023}, pages
  2349--2359.

\bibitem[{Feng et~al.(2020)Feng, Lou, Jiang, and Fox}]{feng2020cryptogru}
Bo~Feng, Qian Lou, Lei Jiang, and Geoffrey~C Fox. 2020.
\newblock Cryptogru: Low latency privacy-preserving text analysis with gru.
\newblock \emph{arXiv preprint arXiv:2010.11796}.

\bibitem[{Guo et~al.(2019)Guo, Mao, and Zhang}]{guo2019augmenting}
Hongyu Guo, Yongyi Mao, and Richong Zhang. 2019.
\newblock Augmenting data with mixup for sentence classification: An empirical
  study.
\newblock \emph{arXiv preprint arXiv:1905.08941}.

\bibitem[{Hoory et~al.(2021)Hoory, Feder, Tendler, Erell, Peled-Cohen, Laish,
  Nakhost, Stemmer, Benjamini, Hassidim, and
  Matias}]{hoory-etal-2021-learning-evaluating}
Shlomo Hoory, Amir Feder, Avichai Tendler, Sofia Erell, Alon Peled-Cohen, Itay
  Laish, Hootan Nakhost, Uri Stemmer, Ayelet Benjamini, Avinatan Hassidim, and
  Yossi Matias. 2021.
\newblock \href {https://doi.org/10.18653/v1/2021.findings-emnlp.102} {Learning
  and evaluating a differentially private pre-trained language model}.
\newblock In \emph{Findings of the Association for Computational Linguistics:
  EMNLP 2021}, pages 1178--1189, Punta Cana, Dominican Republic. Association
  for Computational Linguistics.

\bibitem[{Kahla et~al.(2022)Kahla, Chen, Just, and Jia}]{kahla2022label}
Mostafa Kahla, Si~Chen, Hoang~Anh Just, and Ruoxi Jia. 2022.
\newblock Label-only model inversion attacks via boundary repulsion.
\newblock In \emph{Proceedings of the IEEE/CVF Conference on Computer Vision
  and Pattern Recognition}, pages 15045--15053.

\bibitem[{Kim et~al.(2021)Kim, Choo, Jeong, and Song}]{kim2021co}
JangHyun Kim, Wonho Choo, Hosan Jeong, and Hyun~Oh Song. 2021.
\newblock Co-mixup: Saliency guided joint mixup with supermodular diversity.
\newblock In \emph{International Conference on Learning Representations}.

\bibitem[{Li et~al.(2018)Li, Baldwin, and Cohn}]{li-etal-2018-towards}
Yitong Li, Timothy Baldwin, and Trevor Cohn. 2018.
\newblock \href {https://doi.org/10.18653/v1/P18-2005} {Towards robust and
  privacy-preserving text representations}.
\newblock In \emph{Proceedings of the 56th Annual Meeting of the Association
  for Computational Linguistics (Volume 2: Short Papers)}, pages 25--30,
  Melbourne, Australia. Association for Computational Linguistics.

\bibitem[{Mahmood et~al.(2022)Mahmood, Ahmad, Shafiq, Srinivasan, and
  Zaffar}]{mahmood2022girl}
Asad Mahmood, Faizan Ahmad, Zubair Shafiq, Padmini Srinivasan, and Fareed
  Zaffar. 2022.
\newblock A girl has no name: Automated authorship obfuscation using mutant-x.
\newblock \emph{Proceedings on Privacy Enhancing Technologies}, 1:18.

\bibitem[{Mao et~al.(2011)Mao, Shuai, and Kapadia}]{mao2011loose}
Huina Mao, Xin Shuai, and Apu Kapadia. 2011.
\newblock Loose tweets: an analysis of privacy leaks on twitter.
\newblock In \emph{Proceedings of the 10th annual ACM workshop on Privacy in
  the electronic society}, pages 1--12.

\bibitem[{Pennington et~al.(2014)Pennington, Socher, and
  Manning}]{pennington-etal-2014-glove}
Jeffrey Pennington, Richard Socher, and Christopher Manning. 2014.
\newblock \href {https://doi.org/10.3115/v1/D14-1162} {{G}lo{V}e: Global
  vectors for word representation}.
\newblock In \emph{Proceedings of the 2014 Conference on Empirical Methods in
  Natural Language Processing ({EMNLP})}, pages 1532--1543, Doha, Qatar.
  Association for Computational Linguistics.

\bibitem[{Plant et~al.(2021)Plant, Gkatzia, and
  Giuffrida}]{plant-etal-2021-cape}
Richard Plant, Dimitra Gkatzia, and Valerio Giuffrida. 2021.
\newblock \href {https://doi.org/10.18653/v1/2021.emnlp-main.628} {{CAPE}:
  Context-aware private embeddings for private language learning}.
\newblock In \emph{Proceedings of the 2021 Conference on Empirical Methods in
  Natural Language Processing}, pages 7970--7978, Online and Punta Cana,
  Dominican Republic. Association for Computational Linguistics.

\bibitem[{Qu et~al.(2021)Qu, Kong, Yang, Zhang, Bendersky, and
  Najork}]{qu2021natural}
Chen Qu, Weize Kong, Liu Yang, Mingyang Zhang, Michael Bendersky, and Marc
  Najork. 2021.
\newblock Natural language understanding with privacy-preserving bert.
\newblock In \emph{Proceedings of the 30th ACM International Conference on
  Information \& Knowledge Management}, pages 1488--1497.

\bibitem[{Rajpurkar et~al.(2016)Rajpurkar, Zhang, Lopyrev, and
  Liang}]{rajpurkar-etal-2016-squad}
Pranav Rajpurkar, Jian Zhang, Konstantin Lopyrev, and Percy Liang. 2016.
\newblock \href {https://doi.org/10.18653/v1/D16-1264} {{SQ}u{AD}: 100,000+
  questions for machine comprehension of text}.
\newblock In \emph{Proceedings of the 2016 Conference on Empirical Methods in
  Natural Language Processing}, pages 2383--2392, Austin, Texas. Association
  for Computational Linguistics.

\bibitem[{Sen(2015)}]{sen2015security}
Jaydip Sen. 2015.
\newblock Security and privacy issues in cloud computing.
\newblock In \emph{Cloud technology: concepts, methodologies, tools, and
  applications}, pages 1585--1630. IGI global.

\bibitem[{Shejwalkar et~al.(2021)Shejwalkar, Inan, Houmansadr, and
  Sim}]{shejwalkar2021membership}
Virat Shejwalkar, Huseyin~A Inan, Amir Houmansadr, and Robert Sim. 2021.
\newblock Membership inference attacks against nlp classification models.
\newblock In \emph{NeurIPS 2021 Workshop Privacy in Machine Learning}.

\bibitem[{Socher et~al.(2013)Socher, Perelygin, Wu, Chuang, Manning, Ng, and
  Potts}]{socher-etal-2013-recursive}
Richard Socher, Alex Perelygin, Jean Wu, Jason Chuang, Christopher~D. Manning,
  Andrew Ng, and Christopher Potts. 2013.
\newblock \href {https://aclanthology.org/D13-1170} {Recursive deep models for
  semantic compositionality over a sentiment treebank}.
\newblock In \emph{Proceedings of the 2013 Conference on Empirical Methods in
  Natural Language Processing}, pages 1631--1642, Seattle, Washington, USA.
  Association for Computational Linguistics.

\bibitem[{Song and Raghunathan(2020)}]{song2020information}
Congzheng Song and Ananth Raghunathan. 2020.
\newblock Information leakage in embedding models.
\newblock In \emph{Proceedings of the 2020 ACM SIGSAC conference on computer
  and communications security}, pages 377--390.

\bibitem[{Sun et~al.(2022)Sun, Shao, Qian, Huang, and Qiu}]{sun2022black}
Tianxiang Sun, Yunfan Shao, Hong Qian, Xuanjing Huang, and Xipeng Qiu. 2022.
\newblock Black-box tuning for language-model-as-a-service.
\newblock In \emph{International Conference on Machine Learning}, pages
  20841--20855. PMLR.

\bibitem[{Tang et~al.(2016)Tang, Cui, Li, Ren, Liu, and
  Buyya}]{tang2016ensuring}
Jun Tang, Yong Cui, Qi~Li, Kui Ren, Jiangchuan Liu, and Rajkumar Buyya. 2016.
\newblock Ensuring security and privacy preservation for cloud data services.
\newblock \emph{ACM Computing Surveys (CSUR)}, 49(1):1--39.

\bibitem[{Wang et~al.(2018)Wang, Singh, Michael, Hill, Levy, and
  Bowman}]{wang-etal-2018-glue}
Alex Wang, Amanpreet Singh, Julian Michael, Felix Hill, Omer Levy, and Samuel
  Bowman. 2018.
\newblock \href {https://doi.org/10.18653/v1/W18-5446} {{GLUE}: A multi-task
  benchmark and analysis platform for natural language understanding}.
\newblock In \emph{Proceedings of the 2018 {EMNLP} Workshop {B}lackbox{NLP}:
  Analyzing and Interpreting Neural Networks for {NLP}}, pages 353--355,
  Brussels, Belgium. Association for Computational Linguistics.

\bibitem[{Xiao et~al.(2023)Xiao, Lin, and Han}]{xiao2023offsite}
Guangxuan Xiao, Ji~Lin, and Song Han. 2023.
\newblock Offsite-tuning: Transfer learning without full model.
\newblock \emph{arXiv preprint arXiv:2302.04870}.

\bibitem[{Xu et~al.(2020)Xu, Aggarwal, Feyisetan, and
  Teissier}]{xu-etal-2020-differentially}
Zekun Xu, Abhinav Aggarwal, Oluwaseyi Feyisetan, and Nathanael Teissier. 2020.
\newblock \href {https://doi.org/10.18653/v1/2020.privatenlp-1.2} {A
  differentially private text perturbation method using regularized mahalanobis
  metric}.
\newblock In \emph{Proceedings of the Second Workshop on Privacy in NLP}, pages
  7--17, Online. Association for Computational Linguistics.

\bibitem[{Yue et~al.(2021)Yue, Du, Wang, Li, Sun, and
  Chow}]{yue-etal-2021-differential}
Xiang Yue, Minxin Du, Tianhao Wang, Yaliang Li, Huan Sun, and Sherman S.~M.
  Chow. 2021.
\newblock \href {https://doi.org/10.18653/v1/2021.findings-acl.337}
  {Differential privacy for text analytics via natural text sanitization}.
\newblock In \emph{Findings of the Association for Computational Linguistics:
  ACL-IJCNLP 2021}, pages 3853--3866, Online. Association for Computational
  Linguistics.

\bibitem[{Zhang et~al.(2018)Zhang, Cisse, Dauphin, and
  Lopez-Paz}]{zhang2018mixup}
Hongyi Zhang, Moustapha Cisse, Yann~N Dauphin, and David Lopez-Paz. 2018.
\newblock mixup: Beyond empirical risk minimization.
\newblock In \emph{International Conference on Learning Representations}.

\bibitem[{Zhou et~al.(2022)Zhou, Lu, Gui, Ma, Fei, Wang, Ding, Cheung, Zhang,
  and Huang}]{zhou-etal-2022-textfusion}
Xin Zhou, Jinzhu Lu, Tao Gui, Ruotian Ma, Zichu Fei, Yuran Wang, Yong Ding,
  Yibo Cheung, Qi~Zhang, and Xuanjing Huang. 2022.
\newblock \href {https://aclanthology.org/2022.emnlp-main.572} {{T}ext{F}usion:
  Privacy-preserving pre-trained model inference via token fusion}.
\newblock In \emph{Proceedings of the 2022 Conference on Empirical Methods in
  Natural Language Processing}, pages 8360--8371, Abu Dhabi, United Arab
  Emirates. Association for Computational Linguistics.

\end{thebibliography}
\bibliographystyle{acl_natbib}

\clearpage

\appendix

\section{Privacy Discussion}
\label{sec:privacy-discussion}

In this section, we formally define the threat model in \Cref{sec:threat-model}, and analyze \modelname's privacy from the perspective of possible attacks in \Cref{sec:privacy-analysis}. Moreover, we discuss the incurred cost of privacy in \Cref{sec:privacy-cost}.

\subsection{Threat Model} 
\label{sec:threat-model}

According to \Cref{equ:problem-definition}, we consider a threat model where the adversary $\mathcal{A}$ collects data from the output of $E(\bm{x})$ and $y'$, that is, the input representation and obfuscated decision. Additionally, as a malicious cloud service provider, $\mathcal{A}$ also has white-box access to $M$ and maintains $M$'s training data. $\mathcal{A}$ seeks to reverse the original input text $\bm{x}$ and/or recover the true output $y$.

\subsection{Privacy Analyses}
\label{sec:privacy-analysis}

In \modelname, $E(\bm{x})$ is the PPRG-encoded $[b;x]$ and $b$ pair. In order to acquire $x$, $\mathcal{A}$ needs to recover the text representation of all $[b;x]$ and $b$, as well as pairing them up. As mentioned in \Cref{sec:instance-obfuscation}, a qualified PPRG ensures the irreversibility of the input text sequence, meanwhile generating distinct representations even for identical input. 
Even though $\mathcal{A}$ has training data of $M$ and white-box access to $M$, these do not help $\mathcal{A}$ reverse PPRG-encoded instances as long as PPRG has sufficient privacy strength for resisting known attacks~\cite{song2020information}.
Thus, we conclude that (1) The obfuscator $b$ or the obfuscated instance $[b;x]$ can not be reverse-engineered. (2) Every time the produced representation of $b$ or $[b;x]$ is different, and it is not possible to identify the same $b$ or $[b;x]$ from their PPRG-produced representation. (3) $\mathcal{A}$ is not able to differentiate $[b;x]$ from $b$. (4) No $\bm{x}$ could be extracted from $[b;x]$.


When it comes to resolving a $y$ from $y'$s, as is illustrated in \Cref{equ:decoding-multi-y}, $\mathcal{A}$ has to collect all the associated $[b;x]$ and $b$ pairs in a group $G_n$.
However, identifying a group of $b$s or $[b;x]$s is no better than exhausting all the possible PLM inference request combinations because of the hardness of reversing PPRG-encoded instances and the arbitrary request order of unique representations. For example, let $m=|g|$ be the unit group size and $n$ be the number of unit groups that participated in one instance obfuscation, resolving a $y$ involves $2mn$ independent PLM inference requests. Since $b \in G_n$ is uniformly distributed, the probability of $\mathcal{A}$ to find all these $2mn$ instances in $r$ total requests to the PLM is $1/\binom{r}{2mn}$. Even in a toy setup, say 3-class classification ($m=3$) with one group ($n=1$) of obfuscators, merely 100 total requests would require more than a billion tries to identify all of them.
Therefore, it is impossible for $\mathcal{A}$ to extract $\bm{x}$ from encoded $b$ and $[b;x]$ with a qualified PPRG method and reversing $y$ from $y'$ is no better than exhausting all the possible combinations of requests.

\subsection{Privacy Cost} 
\label{sec:privacy-cost}

Privacy comes with a cost. Here, we elaborate it from two aspects: communication and computation.

\stitle{Communication cost} As a baseline, each instance $x$ sends one request to the PLM. In \modelname, as \Cref{equ:obfuscator-group}, each $x$ is concatenated with $|G_n|=n|C|$ instances. All these obfuscated instances, along with the same amount of obfuscators, form the total requests to the PLM, namely, $2n|C|$. In practice, when there are multiple $x$s, the obfuscators are pre-computed and reused from the obfuscator pool. Hence, for $k$ $x$ instances, the total number of requests ranges in $[(1+kn)|C|, 2kn|C|]$.

\stitle{Computation cost} On PLM's side, the number of requests to inference is indicated in the communication cost. On the data owner's side, resolving a $y$ requires the execution of \Cref{equ:decision-resolution}, which only involves trivial matrix operations.

\section{Experiments}

We report additional experiments and analyses in this section.

\subsection{Length Expansion}
\label{sec:length-expansion}

The privacy strength of a single obfuscated instance $[b;x]$ mainly comes from the domination of $b$. \citet{kim2021co} demonstrate that the model has the ability to map a mixed instance consisting of more than two raw instances to a mixed label. Inspired by that, We expand the length of $b$ to amplify the impact of it on PLM's decision by duplicating it $k$ times.~\footnote{The duplication does not hurt the privacy of the generated representation because it is before semantic-neutral shuffling.} Correspondingly, the concatenation sequence length becomes $k|b|+|x|$. Here, we seek to verify the relation between the accuracy and the obfuscator's length over obfuscated and resolved prediction results.

\begin{figure}[!t]
    \centering
    \includegraphics[width=1\linewidth]{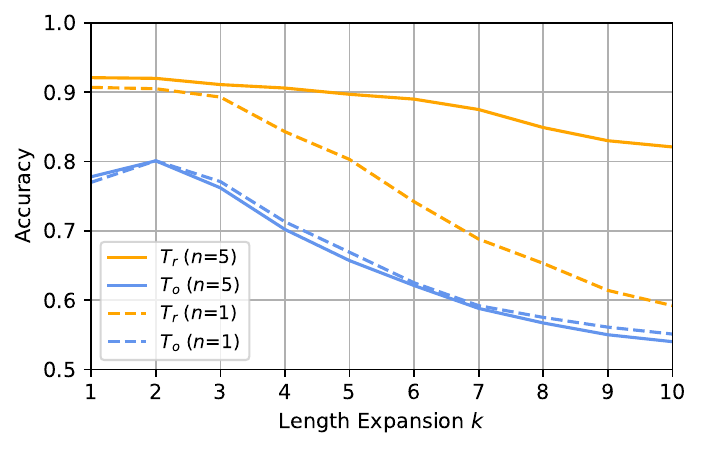}
    \vspace{-2em}
    \caption{Length Expansion (SST-2)}
    \label{fig:exp-length-expansion}
    \vspace{-1.5em}
\end{figure}

We duplicate $b$ by $k$ times before encoding to realize the length increment, and the dataset we used here is SST-2. As the solid lines shown in \Cref{fig:exp-length-expansion}, $k$ is tested from 1 to 10 consecutively, and it presents a negative correlation to accuracy. Specifically, when $k=1$, the accuracy of resolved inference is more than 0.9 whereas the accuracy of obfuscated inference is less than 0.8. When $k$ becomes larger, the accuracy of resolved inference $T_r$ drops gradually until it reaches around 0.82 when $k$ is 10. As a comparison, the trend of obfuscated $T_o$ falls quickly almost throughout the $k$'s process and when $k$ is 10, it hits 0.55, which is close to the random guess. The maximum difference of the accuracy between two variants is more than 2.25 times to the minimum difference at the beginning ($k$=2). This experiment demonstrates the massive impact of length expansion on the performance, and a proper $k$ could deviate the obfuscated distribution of PLM's decision far from the true one, thus intensifying the privacy protection. 
Note that we set another hyper-parameter $n$ in $G_n$ to be 1 and 5 in this experiment; regardless of $n$, the negative correlation holds. 


\subsection{Decision Distribution}

According to \Cref{equ:decision-privacy-distribution}, the decision distribution of $M(b)$ or $M([b;x])$ should be as close to random as possible, and the overall decision distribution of $M(\cdot)$ should also lean towards randomness.

For validation purposes, we conduct an additional experiment in \Cref{tab:decision-distribution-sst2}, which presents the decision distribution of $M(b)$, $M([b;x])$ and the overall $M(\cdot)$ on SST-2. Specifically, each cell denotes the distribution of negative/positive decisions. The parameter $k$ controls the strength of (as introduced in \Cref{sec:length-expansion}, a higher $k$ enhances privacy but reduces utility). 

At $k=1$ (optimal utility), the distributions of $M(b)$, $M([b;x])$ and $M(\cdot)$ approach randomness. As $k$ increases, the level of randomization intensifies. At $k=10$, the distribution becomes equivalent to random.

\begin{table}[t]
\centering
\small
\setlength{\tabcolsep}{5pt}
\begin{tabular}{l|ccc}
\hline
$k$ & $M([b;x])$ & $M(b)$ & $M(\cdot)$ \\
\hline
$1$ & 0.431/0.569 & 0.5/0.5 & 0.465/0.535 \\
\rowcolor{gray!20}$5$ & 0.487/0.513	& 0.5/0.5 & 0.493/0.507 \\
$10$ & 0.502/0.498 & 0.5/0.5 & 0.501/0.499 \\
\hline
\end{tabular}
\caption{Decision distribution (SST-2)}
\label{tab:decision-distribution-sst2}
\end{table}

Moreover, we set $k=1$, and report the decision distribution for the other three datasets in \Cref{tab:decision-distribution-others}. Similarly, each cell denotes the distribution of decisions. The results show that our method achieves a promising decision distribution even with optimal utility.

\begin{table}[t]
\centering
\small
\setlength{\tabcolsep}{5pt}
\begin{tabular}{l|ccc}
\hline
Dataset & $M([b;x])$ & $M(b)$ & $M(\cdot)$ \\
\hline
SST-5 & \makecell{0.141/0.227/0.191\\0.273/0.168} & \makecell{0.2/0.2/0.2\\0.2/0.2} & \makecell{0.171/0.213/0.195\\0.236/0.184} \\
\rowcolor{gray!20}MRPC & 0.528/0.472 & 0.5/0.5 & 0.514/0.486 \\
QNLI & 0.532/0.468 & 0.5/0.5 & 0.516/0.484 \\
\hline
\end{tabular}
\caption{Decision distribution ($k=1$)}
\label{tab:decision-distribution-others}
\end{table}

\end{document}